\documentclass{article}
\usepackage{iclr2027_conference,times}
\iclrfinalcopy
\usepackage{amsmath,amssymb,amsthm,booktabs,array,tabularx,multirow}
\usepackage{graphicx,xcolor,microtype}
\usepackage{placeins}
\usepackage{hyperref}
\hypersetup{colorlinks=true,linkcolor=black,citecolor=black,urlcolor=blue}
\newcommand{\E}{\mathbb{E}}

\newcommand{\given}{\,\vert\,}
\newcolumntype{Y}{>{\raggedright\arraybackslash}X}
\providecommand{\propositionname}{Proposition}

\newcommand{\papertitle}{Game-Enhanced Embodied Learning with Event and Foresight Pretraining}

\newcommand{\paperauthorsmetadata}{Linhao Zhou, Lianlei Shan, Han Jin, Yaunchao Ma, Xinpeng Chen, Binxin Liu, Bihao Wang}
\author{%
Linhao Zhou\textsuperscript{1}\thanks{Equal contribution.},\quad
Lianlei Shan\textsuperscript{2}\footnotemark[1],\quad
Han Jin\textsuperscript{3}\thanks{Corresponding authors. General contact: Lianlei Shan
(\href{mailto:shanlianlei18@mails.ucas.edu.cn}{\texttt{shanlianlei18@mails.ucas.edu.cn}}).},\quad
Yaunchao Ma\textsuperscript{3}\footnotemark[2]\\
\bfseries Xinpeng Chen\textsuperscript{4},\quad
Binxin Liu\textsuperscript{3}\footnotemark[2],\quad
Bihao Wang\textsuperscript{3}\\
\normalfont
\textsuperscript{1}Peking University \quad
\textsuperscript{2}Tsinghua University\\
\textsuperscript{3}Hakko AI PTE. LTD. \quad
\textsuperscript{4}University of Chinese Academy of Sciences%
}

\title{\papertitle}
\date{}
\hypersetup{pdftitle={\papertitle},pdfauthor={\paperauthorsmetadata}}
\begin{document}
\maketitle
\fancyhead{}
\renewcommand{\headrulewidth}{0pt}
\begin{abstract}
Human gameplay supplies paired observations, actions, and visual consequences that can supplement robot-policy training. We study event-and-foresight pretraining for InternVLA-A1.5, combining recorded keyboard/mouse event prediction with history-conditioned visual supervision through a frozen video model. Coverage-preserving anchor selection constructs one-second targets, and an attention mask prevents teacher-forced future actions from entering foresight queries. The selected corpus contains 394 recordings across 20 games, with 223.36 hours of recorded timestamp span and 2,072,084 training anchors. After 12,400 game-pretraining updates, we compare the resulting initialization with the public base under the same suite-specific robot fine-tuning protocol. Across four LIBERO suites, aggregate success increases from 1,718/2,000 (85.90\%) to 1,811/2,000 (90.55\%), a gain of 4.65 percentage points. Suite gains are 9.2, 6.8, 0.2, and 2.4 points on Spatial, Object, Goal, and Long. The 40-task breakdown shows 24 improvements, 10 declines, and six ties. These results support the practical value of the complete game-pretraining stage in this evaluated setting, while revealing heterogeneous task outcomes. The comparison includes additional optimization and does not isolate the effects of game data, the two objectives, or training variability.
\end{abstract}

\section{Introduction}
Embodied policies learn from observations paired with actions and subsequent environmental changes. Human gameplay provides the same broad interaction structure: a player observes a scene, issues keyboard or mouse commands, and observes their consequences. These recordings are a potential source of auxiliary supervision for robot-capable models. Our question is how to train on that non-robot action space while retaining the policy's native robot interface, and what downstream changes accompany the additional training stage.

D2E demonstrates that desktop interaction can support embodied pretraining \citep{choi2026d2e}, while InternVLA-A1.5 provides a robot-capable architecture with learnable foresight queries supervised by a frozen video model \citep{ma2026internvla}. Adapting game interaction to this architecture is not direct: game actions are expressed as keyboard and mouse events rather than robot controls, action and video targets must be aligned over physical time, and teacher-forced future events must not leak into features available at inference. The central challenge is therefore to extract auxiliary supervision from game interaction while maintaining causal, history-only conditioning and preserving the model's native robot-control pathway.

To address these constraints, our procedure builds one-second action and video targets from recorded human gameplay. Coverage-preserving sampling combines a sparse-anchor pass that reduces repeated starting points with a recovery pass that restores valid anchors for otherwise uncovered actions. Next-event prediction learns from recorded human commands, while the foresight pathway receives future visual supervision conditioned only on available history. Target-action tokens cannot enter the foresight queries. The trained components subsequently initialize native robot fine-tuning.

To assess transfer, we compare the game-pretrained initialization with the public base across all four LIBERO suites, holding suite-specific robot training and rollout protocols fixed. The game-pretrained initialization produces 93 additional successes in 2,000 episodes, increasing aggregate success from 85.90\% to 90.55\%. Spatial and Object account for 80 of these 93 net additional successes. Full task-level results show both improvements and regressions, so the aggregate benefit is not interpreted as a uniform task-level effect.

Taken together, our contribution is a concrete training route for using non-robot interaction as auxiliary supervision without converting game commands into robot actions. The route combines coverage-preserving game sampling, causal event-and-foresight training, transfer to InternVLA-A1.5's native robot pathway, and a complete four-suite evaluation. Table~\ref{tab:positioning} identifies which components are inherited from D2E and InternVLA-A1.5 and which adaptations are evaluated in this study.

\section{Related Work}
\paragraph{Learning from game interaction.}
VPT uses inverse dynamics to label gameplay videos \citep{baker2022vpt}; SIMA studies instruction following across virtual environments \citep{sima2024}; Gato combines heterogeneous tasks in a sequence model \citep{reed2022gato}. D2E develops a desktop-data pipeline and demonstrates embodied transfer through vision--action pretraining \citep{choi2026d2e}. We adopt that data direction and study a concrete event-and-foresight adaptation of a robot-capable model. Table~\ref{tab:positioning} identifies the inherited ingredients and this study's contribution.

\paragraph{Robot policies and visual foresight.}
OpenVLA adapts multimodal representations to robot control \citep{kim2024openvla}, and FAST studies tokenization of continuous actions \citep{pertsch2025fast}. InternVLA-A1.5 combines understanding, latent foresight, and action prediction \citep{ma2026internvla}. We retain its robot architecture and extend its training supervision to human game events and their future visual context. Frozen Wan2.2 TI2V-5B modules provide the visual objective \citep{wan2025,wan22}, and LIBERO supplies the downstream manipulation tasks \citep{liu2023libero}.

\paragraph{Objectives and interaction representations.}
Flow matching supplies the continuous video-supervision framework \citep{lipman2023flow}. LoRA provides the low-rank updates used in our supporting inverse-dynamics diagnostics \citep{hu2022lora}. These diagnostics examine event reconstruction separately from robot transfer. Their role is to characterize the quality of structured interaction targets; human event labels define the evaluated game-pretraining route.

\begin{table}[!htbp]
\caption{Relationship to the two closest methods. Our contribution is the specified game-supervision and transfer procedure; the underlying event format and foresight architecture retain their original attribution.}
\label{tab:positioning}\centering\small
\begin{tabularx}{\linewidth}{p{.19\linewidth}YY}\toprule
Work & Established ingredient & Adaptation evaluated here\\\midrule
D2E \citep{choi2026d2e} & Desktop events and vision--action pretraining for embodied transfer & Recorded human events paired with future game frames for joint event-and-foresight training\\
InternVLA-A1.5 \citep{ma2026internvla} & Native robot expert, foresight queries, and frozen-video supervision & Causal game-event targets, history-only foresight conditioning, and transfer back to the robot path\\
This study & Both ingredients above & Coverage-preserving game samples, a joint training interface, and a fixed downstream initialization comparison\\\bottomrule
\end{tabularx}
\end{table}

\section{Game Event-and-Foresight Pretraining}\label{sec:method}
\subsection{Game interaction as auxiliary embodied supervision}
\begin{figure}[t]
\centering
\includegraphics[width=\linewidth]{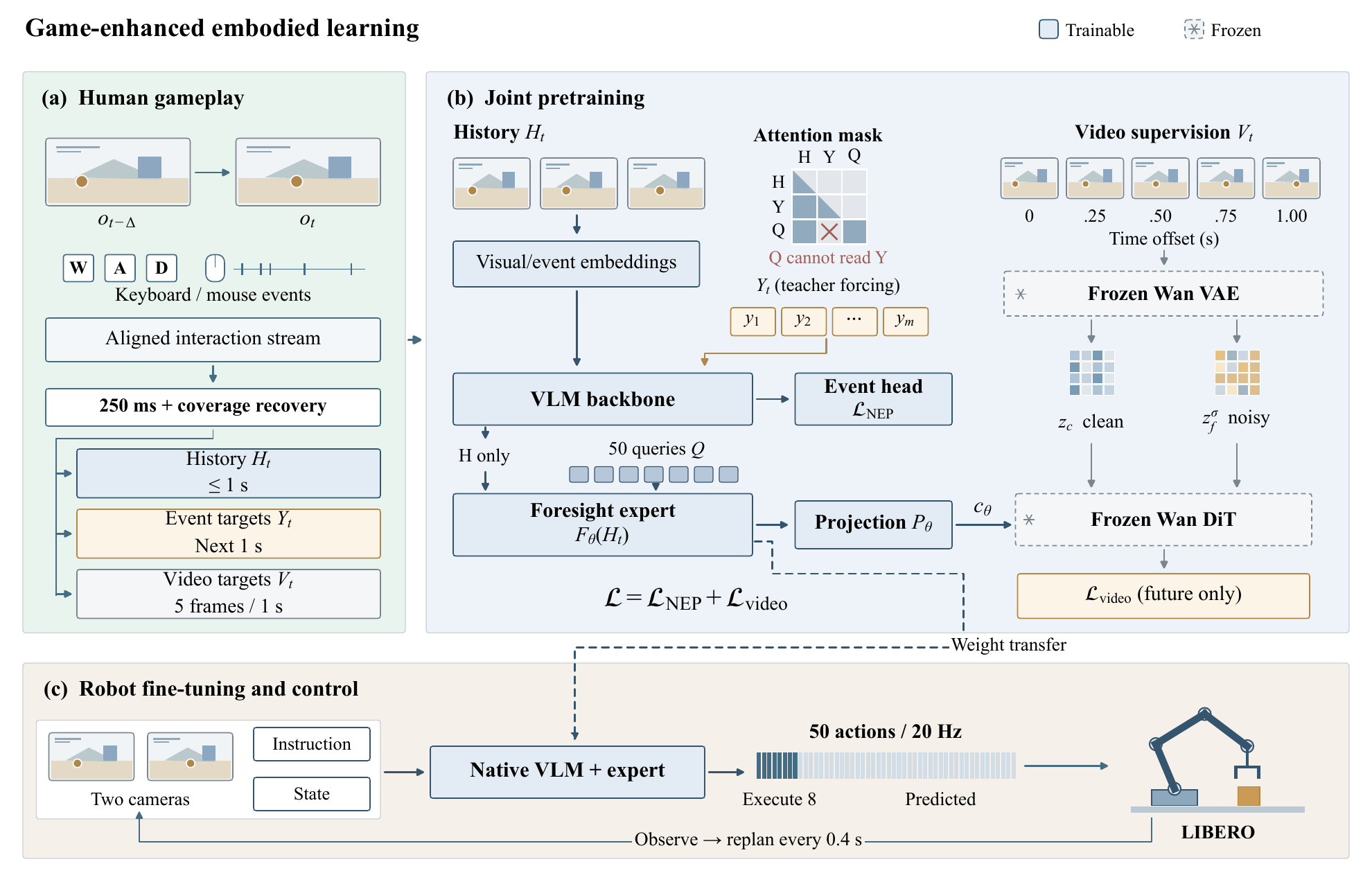}
\caption{Game-enhanced embodied training. (a) Recorded human gameplay supplies recent history $H$, next-action tokens $Y$, and future video labels. (b) Event prediction and visual foresight jointly train the multimodal model. In the attention inset, rows are query groups and columns are key groups; $Q$ denotes foresight queries, triangles denote causal access, and $Q$ cannot access $Y$. Frozen video modules supervise the future latent and pass conditioning gradients to trainable modules. (c) Game-pretrained weights initialize the native robot policy, which predicts 50 actions and executes eight before replanning. Screens and the robot are schematic.}
\label{fig:overview}\label{fig:method}
\end{figure}

Let a human game recording be $\mathcal D_G=\{e_i\}_{i=1}^{N}$, with event $e_i=(c_i,t_i,u_i)$ containing its type, physical timestamp, and payload. Screen payloads contain images; action payloads contain keyboard or mouse input. We retain the recorded event ordering and timestamps when constructing causal samples. Figure~\ref{fig:method} summarizes the training route from game records to robot execution.

Starting from the public model parameters $\theta_0$, the procedure is
\begin{equation}
\theta_G=\operatorname{Pretrain}(\theta_0,\mathcal D_G),\qquad
\theta_R=\operatorname{Finetune}(\operatorname{Bridge}(\theta_G),\mathcal D_R),
\label{eq:transfer}
\end{equation}
where $\mathcal D_R$ contains robot demonstrations. The bridge preserves learned weights and selects the native robot forward path. Thus game data supplies additional parameter training before robot fine-tuning; keyboard and mouse commands are not treated as robot motor targets. The comparison arm fine-tunes directly from the public robot-capable base on the same $\mathcal D_R$.

The game branch uses the expanded 250,416-row vocabulary to predict serialized event tokens; its continuous robot-action objective is disabled. The foresight queries and projection are trained jointly with the multimodal backbone and expert. At transfer, the bridge consolidates the saved tensors and restores the native robot forward path. It does not learn an additional mapping from keyboard or mouse commands to motor actions.

\subsection{Causal samples and attention masks}
Each game-interaction sample uses a screen anchor $t$, a recent history $H_t$ of at most one second, and target action-event tokens $Y_t$ from $(t,t+1\,\mathrm{s}]$. Anchors follow a 250 ms cadence, with additional eligible anchors covering otherwise uncovered actions. Future screens serve as video labels and are excluded from the language-model input. With batch size $B$ and target-token set $\mathcal T_b$, the event objective is
\begin{equation}
\mathcal L_{\mathrm{NEP}}=-\frac1B\sum_{b=1}^{B}\frac1{|\mathcal T_b|}\sum_{j\in\mathcal T_b}\log p_\theta(y_{bj}\given H_{t_b},y_{b,<j}).
\label{eq:nep}
\end{equation}
This reduction gives each sample equal weight after averaging over its target tokens.

Fifty learnable foresight queries produce $F_\theta(H_t)$. Their attention to every target-action position is blocked, and their position indices are anchored before the target suffix. Thus teacher-forced future actions cannot directly enter the foresight features. The event head learns the causal action sequence, while the foresight pathway learns from history-conditioned future visual supervision.

\subsection{Frozen video supervision}
We sample video frames at offsets $[0,0.25,0.50,0.75,1.00]$ seconds. A frozen Wan VAE maps the five-frame clip into two temporal latent slices; a separately encoded current image provides the clean condition slice. Let $z_f$ denote the future slice, $\epsilon$ sampled noise, and $\sigma$ the scheduler noise level. The frozen denoiser receives
\begin{equation}
\widetilde z_\sigma=[z_c,(1-\sigma)z_f+\sigma\epsilon],\qquad c_\theta=P_\theta F_\theta(H_t).
\end{equation}
The condition patch has timestep zero and the future patch has the sampled timestep. With $M$ masking out the condition slice, the video loss is
\begin{equation}
\mathcal L_{\mathrm{video}}=\E\!\left[\frac{\left\|M\odot\big(v_\omega(\widetilde z_\sigma,c_\theta,\boldsymbol\sigma)-v^\star\big)\right\|_2^2}{N_z}\right],\qquad v^\star=[0,\epsilon-z_f].
\label{eq:video}
\end{equation}
The denominator $N_z$ counts all latent elements, including the masked condition slice. With two equally sized temporal slices, this is one half of the future-slice mean squared error. Noise is standard Gaussian; a uniform index selects a level from the 1,000-entry, shift-5.0 scheduler (Appendix~\ref{app:loss-conventions}). Unit coefficients do not imply equal gradient contributions. The VAE and denoiser weights $\omega$ remain frozen; gradients through conditioning train the foresight pathway. We optimize $\mathcal L=\mathcal L_{\mathrm{NEP}}+\mathcal L_{\mathrm{video}}$, with unit coefficients. The trainable modules are the visual encoder, multimodal backbone, expert, foresight queries, and projection.

\subsection{Checkpoint transfer and supervision horizon}
Game-interaction pretraining uses 2,072,084 anchors, global batch $8\times1\times16=128$, AdamW learning rate $2\times10^{-5}$, 200 warmup updates, and seed 42. We evaluate rolling checkpoint 12,400, corresponding to 1,587,200 sample presentations, or 76.6\% of a nominal epoch. The bridge preserves the trained weights while selecting the robot forward path; Appendix~\ref{app:training} details its configuration.

Robot fine-tuning uses the native objective $10\mathcal L_{\mathrm{action}}+\mathcal L_{\mathrm{language}}+\mathcal L_{\mathrm{video}}$. The language term uses the labels available in each robot batch. At 20 Hz, the 50-step action target has offsets $0,0.05,\ldots,2.45$ seconds. The five-frame video rule $k_i=\lfloor50i/4\rfloor$ yields frame offsets $[0,12,25,37,50]$, or $[0,0.60,1.25,1.85,2.50]$ seconds. Figure~\ref{fig:horizon} makes the cross-stage change explicit: both stages use five frames, but the downstream video target spans 2.5 times the physical duration. Action execution uses a separate replanning interval of eight steps, or 0.4 seconds. Frame count, prediction horizon, and execution interval therefore specify different aspects of the policy. This is a cross-stage specification comparison; the initialization experiment holds the downstream horizon fixed and does not estimate the effect of changing that horizon.

\section{Experiments: Game-Pretrained Initialization}\label{sec:experiments}
\subsection{Human game data and sampling}\label{sec:game-data}
We use a selected human-recorded subset of D2E with 20 games, 394 recordings, and 22,646,179 screen, keyboard, and raw-mouse events. The summed per-recording timestamp span is 223.36 hours; the indexed active intervals total 197.85 hours. Neither duration is inferred from overlapping training windows. The data include shooting, driving, building, and other interaction settings; Appendix~\ref{app:game-corpus} lists every game with its counts and durations. Supervision uses original human events.

\begin{table}[!htbp]
\caption{Selected game corpus and sample construction. A recording, an event row, and an overlapping anchor sample are distinct units. Durations sum recording spans and indexed active intervals, respectively.}
\label{tab:game-data}\centering\small
\begin{tabularx}{\linewidth}{Yr}\toprule
Quantity & Count / duration\\\midrule
Games / recorded episodes & 20 / 394\\
Recorded span / active intervals (hours) & 223.36 / 197.85\\
Raw event rows & 22,646,179\\
Legal dense samples after validity checks & 11,118,648\\
Samples after 250 ms anchor selection & 2,356,278\\
Valid anchors restored for action coverage & 3,038\\
Final train / validation anchors & 2,072,084 / 287,232\\\bottomrule
\end{tabularx}
\end{table}

First, eligible screen anchors define at most one second of causal history and a one-second action target. Boundary, corruption, token-budget, and frame-retrieval checks reject invalid candidates. Of 11,797,800 candidates reaching the action-target check, 677,326 (5.74\%) contain no keyboard or mouse event and are removed. Second, greedy 250 ms anchor selection reduces repeated starting points. Third, coverage recovery restores 3,038 legal anchors for 6,681 otherwise uncovered eligible action ordinals, raising coverage from 99.94\% to 100\%. It preserves recorded payloads and timestamps; local anchor gaps can be shorter than 250 ms. Figure~\ref{fig:sampling-coverage} details these counts.

The nominal 90/10 split groups recordings by source session, stratified by game. It assigns 349 recordings in 171 session groups to training and 45 recordings in 23 groups to validation; overlapping windows remain together. The index checks report no overlap in sessions, recordings, media files, or recording--video-timestamp pairs between the partitions. This is a within-game split, not a held-out-game evaluation. Training uses stored epoch permutations without augmentation. Across all final samples, median history length is 32 events including 17 screens, median history span is 968.32 ms, and median active token count is 1,726. The training-only token median is 1,723; the maximum is 4,686, below the 8,192-position limit.

\begin{figure}[!htb]
\centering
\includegraphics[width=\linewidth]{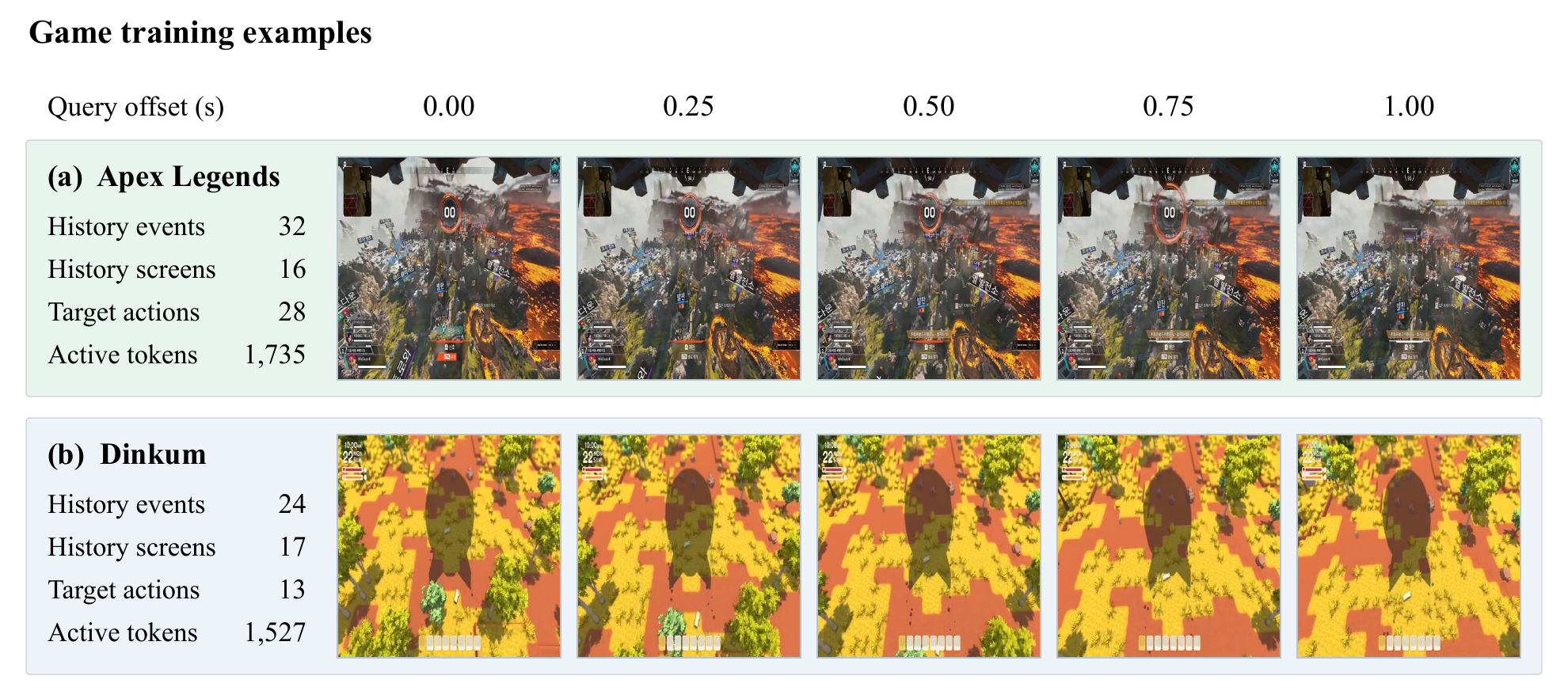}
\caption{Two recorded training examples from the formal index. History and target event counts come from the corresponding sample metadata; the images show the five video-supervision frames. Column times are requested offsets, while actual frames use the preceding available presentation timestamp. The examples are training labels rather than model-generated forecasts.}
\label{fig:real-game-samples}
\end{figure}

\subsection{Comparison and evaluation protocol}
The primary comparison changes the initialization entering robot fine-tuning. The Base arm starts from public InternVLA-A1.5-base; the game-pretrained arm starts from the transferred step-12,400 checkpoint. Each initialization is fine-tuned separately on LIBERO Spatial, Object, Goal, and Long for 20,000 updates per suite. Within each suite, both arms share the robot demonstrations, training code, hyperparameters, and evaluation protocol. Each initialization therefore yields four suite-specific policies. We analyze all four suite scores, their equal-weight aggregate, and all 40 task-level contrasts.

\begin{table}[!htbp]
\caption{The compared training routes. The downstream robot stage is shared; the game-pretrained arm has an additional human-interaction training stage.}
\label{tab:training-routes}\centering\small
\begin{tabularx}{\linewidth}{lYY}\toprule
Stage & Direct Base & Game-pretrained initialization\\\midrule
Starting model & InternVLA-A1.5-base & InternVLA-A1.5-base\\
Game supervision & None & Human event tokens and future frames\\
Additional updates & 0 & 12,400, global batch 128\\
Robot objective & Native action, language, and video losses & Same downstream objectives\\
Robot updates & 20,000 per suite & 20,000 per suite\\
Evaluation & 500 episodes per suite & Same evaluation protocol\\\bottomrule
\end{tabularx}
\end{table}

Each suite-specific policy is evaluated on 10 tasks with 50 episodes per task. Both arms use evaluation seed 7, 10 settling steps, images resized to 224, the standard inference backend, and replanning after eight actions. Internal gripper values in $[0,1]$ are mapped to LIBERO's signed convention. Both arms use the same data conversion and video implementation, including the clean condition latent, zero condition timestep, and sampled future timestep. Each suite uses a batch of 32, a learning rate from $5\times10^{-5}$ to $5\times10^{-6}$ with 400 effective warmup updates, and the step-20,000 checkpoint. Appendix~\ref{app:robot-protocol} specifies optimizer settings, trainable modules, and the distinction between the single-suite and earlier joint-training recipes.

\subsection{Four-suite transfer results}
\begin{figure}[!htbp]
\centering
\includegraphics[width=\linewidth]{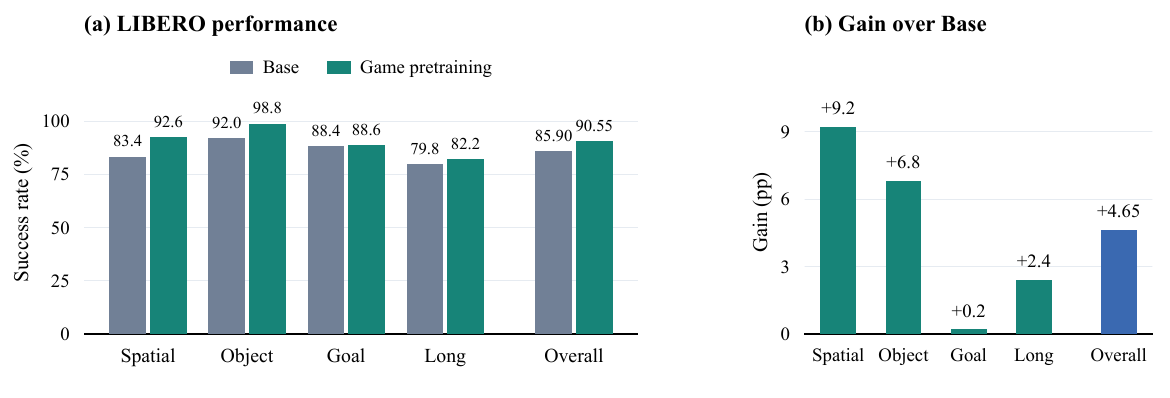}
\caption{Four-suite success rates and differences. The largest observed increases occur on Spatial and Object. Overall is an equal-suite aggregate; all bars are descriptive point estimates under the shared downstream protocol.}
\label{fig:result-libero}
\end{figure}

Table~\ref{tab:libero} and Figure~\ref{fig:result-libero} report all four suites. Game initialization yields 1,811 successes versus 1,718 for Base over 2,000 episodes per arm: 90.55\% versus 85.90\%, a gain of 4.65 percentage points. Since every suite contains 500 episodes, the pooled rate also equals the equal-weight mean of the four suite rates. Each arm comprises four separately fine-tuned policies; the aggregate is not a single joint-policy score.

\begin{table}[!htb]
\caption{Complete LIBERO initialization comparison. Each named suite uses 500 episodes per arm. Parentheses give success rates; differences are game initialization minus Base. Overall pools four suite-specific policies per arm.}
\label{tab:libero}\centering\small
\begin{tabularx}{\linewidth}{lYYrr}\toprule
Suite & Base & Game initialization & $\Delta n$ & $\Delta$ pp\\\midrule
Spatial & 417 (83.4\%) & 463 (92.6\%) & +46 & +9.2\\
Object & 460 (92.0\%) & 494 (98.8\%) & +34 & +6.8\\
Goal & 442 (88.4\%) & 443 (88.6\%) & +1 & +0.2\\
Long & 399 (79.8\%) & 411 (82.2\%) & +12 & +2.4\\\midrule
Overall / 2,000 & 1,718 (85.90\%) & 1,811 (90.55\%) & +93 & +4.65\\\bottomrule
\end{tabularx}
\end{table}

Spatial increases by 9.2 points and Object by 6.8 points. Together they account for 80 of the 93 net additional successes. Long increases by 2.4 points, while Goal changes by one episode (+0.2 points).

\subsection{Task-level heterogeneity}
\begin{figure}[!htbp]
\centering
\includegraphics[width=\linewidth]{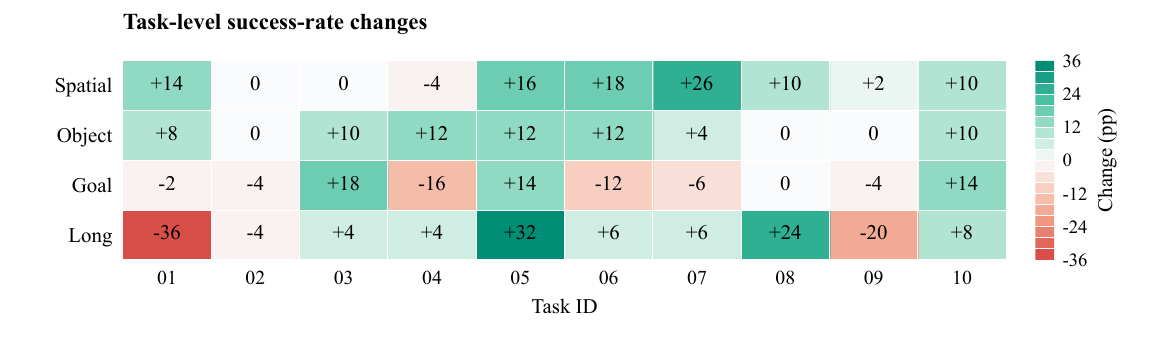}
\caption{Task-level changes in percentage points, game initialization minus Base. Columns index tasks within each suite; the full mapping appears in Appendix~\ref{app:libero-tasks}. These are complete benchmark-task success rates, not completion rates of intermediate steps.}
\label{fig:task-differences}
\end{figure}

Figure~\ref{fig:task-differences} displays all 40 task contrasts; Appendix~\ref{app:libero-tasks} gives their success counts and original task names. Each task has 50 episodes per arm. Success increases on 24 tasks, decreases on 10, and is unchanged on six. Positive suite averages therefore coexist with individual regressions.

On Goal, three tasks improve, six decline, and one ties, leaving a net gain of one success. On Long, placing two mugs on their respective plates improves from 14/50 to 30/50 (+32 points), while placing alphabet soup and tomato sauce in a basket falls from 42/50 to 24/50 (-36 points). These observed contrasts locate the variation but do not identify its mechanism. They also preclude interpreting the suite-level Long gain as uniform improvement on long-horizon tasks.

The two arms have equal downstream budgets but different total training budgets. Their difference evaluates the complete event-and-foresight game stage, including additional optimization; it does not separate the objective terms or isolate game data from compute. Scores from published joint-training policies use a different training recipe and are not treated as a matched comparison.

\FloatBarrier
\section{Interpretation and Scope}\label{sec:analysis}
\paragraph{What is compared.}
The experiment evaluates adding the full event-and-foresight game stage before a fixed robot-training procedure. Its aggregate downstream outcome is +93/2,000 successes, with positive differences in all four suite averages and mixed task-level outcomes. Event targets and future frames provide complementary supervision by design, but their separate contributions are not measured. Both arms use the same robot horizon and local video implementation. Differences in pretraining versus robot horizons are therefore implementation specifications, not a tested explanation of the gains.

\paragraph{Uncertainty and generalization.}
The unit of the reported result is a fixed suite-specific policy evaluated over 10 tasks with 50 episodes each. Repeated rollouts are not independent training replications. The per-task success counts do not reveal which paired initial states changed outcome, so they do not determine an episode-paired significance test; we report no significance or equivalence claim. In particular, the one-episode Goal difference should be interpreted as a small observed change. The results concern the selected human corpus, model, and simulator protocol; they do not establish consistent improvement across all tasks, games, architectures, or real robots. The additional training budget also prevents assigning the difference to game data alone.

\paragraph{Representation diagnostics.}
Appendices~\ref{app:diagnostics} and~\ref{app:cases} retain independent studies of game-event structure, mouse motion, and event density. They motivate careful interaction-label evaluation. Their reconstructed actions do not enter robot pretraining, so these diagnostics are not ablations of the event-and-foresight transfer procedure.

\section{Conclusion}\label{sec:conclusion}
We specify how human game interaction can supplement a robot-capable model through joint event prediction and history-conditioned visual foresight, followed by native robot fine-tuning. In the InternVLA-A1.5 comparison, aggregate LIBERO success rises from 85.90\% to 90.55\% under the shared downstream protocol. Spatial and Object contribute most of the net gain; the complete task breakdown includes both improvements and regressions. These setting-specific observations identify the complete game-pretraining stage as a promising source of auxiliary supervision for the evaluated embodied model.

\section*{Reproducibility Statement}
We specify the event representations, attention masks, loss reductions, training configuration, checkpoint bridge, and evaluation protocol in the paper and appendices. The manuscript package contains the English source and compiled PDF, numerical tables, a source-linked record of dataset and robot configuration values, the transcribed qualitative-case bins, arithmetic checks, frozen figure assets, and archived figure-generation code. The token-availability analysis specifies the training and cached-generation conventions separately. These materials reproduce the reported tabulations and visualizations.

\section*{Ethics Statement}
This study uses existing desktop interaction and robot benchmark data. Handling or releasing desktop recordings requires respecting source permissions and removing identifying information. Robot evaluation is conducted in simulation; the reported metrics measure benchmark task success under the stated protocol. The accompanying materials omit private server paths and identifying run names.

\bibliography{references}
\bibliographystyle{iclr2027_conference}

\clearpage
\appendix
\section{Game Corpus and Sample Construction}\label{app:game-corpus}

\begin{figure}[!htbp]\centering

\includegraphics[width=\linewidth]{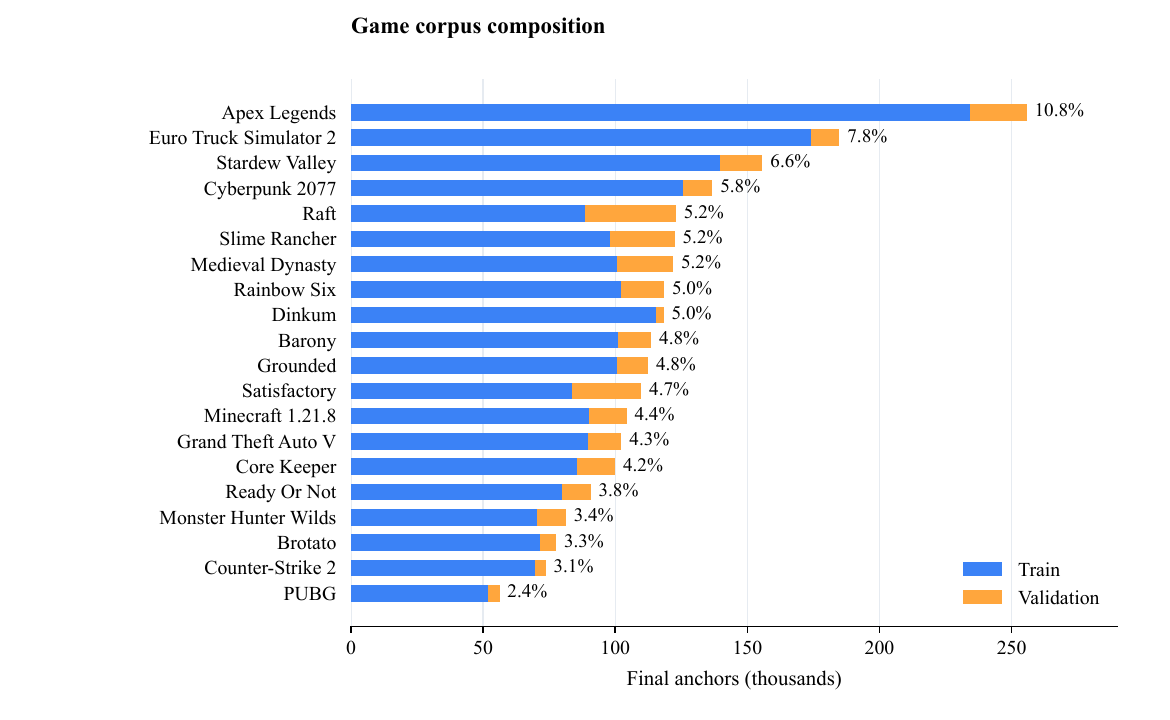}

\caption{Per-game composition of train and validation anchors. Percentages use all final anchors and do not count independent trajectories.}\label{fig:game-composition}\end{figure}

Table~\ref{tab:game-composition} lists the complete selected game roster. Recorded span sums the timestamp range of each MCAP recording; active hours sum the indexed active intervals, giving 223.36 and 197.85 hours, respectively. Figure~\ref{fig:game-composition} reports anchor proportions, a different quantity from duration proportions.

\begin{table}[!htbp]\centering\footnotesize

\caption{Per-game corpus statistics. Durations are in hours; anchors are split by source session.}\label{tab:game-composition}

\begin{tabularx}{\linewidth}{Yrrrrr}\toprule

Game & Records & Span & Active & Train anchors & Val. anchors\\\midrule

Apex Legends & 36 & 24.95 & 20.11 & 234,160 & 21,541\\

Barony & 11 & 9.25 & 9.20 & 101,077 & 12,558\\

Brotato & 13 & 5.93 & 5.81 & 71,558 & 6,057\\

Core Keeper & 19 & 8.87 & 8.13 & 85,467 & 14,486\\

Counter-Strike 2 & 10 & 9.85 & 6.81 & 69,633 & 4,041\\

Cyberpunk 2077 & 33 & 13.98 & 11.95 & 125,783 & 10,992\\

Dinkum & 33 & 10.21 & 9.77 & 115,292 & 3,043\\

Euro Truck Simulator 2 & 35 & 19.48 & 17.78 & 174,104 & 10,740\\

Grand Theft Auto V & 11 & 11.25 & 9.01 & 89,771 & 12,590\\

Grounded & 31 & 9.35 & 9.12 & 100,638 & 11,794\\

Medieval Dynasty & 12 & 10.86 & 10.25 & 100,735 & 21,162\\

Minecraft 1.21.8 & 10 & 8.60 & 8.47 & 89,978 & 14,332\\

Monster Hunter Wilds & 34 & 7.80 & 6.89 & 70,321 & 10,981\\

PUBG & 13 & 4.81 & 4.58 & 51,747 & 4,461\\

Raft & 21 & 10.20 & 10.02 & 88,598 & 34,406\\

Rainbow Six & 11 & 13.69 & 10.02 & 102,099 & 16,393\\

Ready Or Not & 16 & 9.53 & 7.45 & 79,788 & 10,916\\

Satisfactory & 22 & 9.64 & 9.05 & 83,697 & 26,158\\

Slime Rancher & 10 & 10.66 & 10.35 & 98,013 & 24,489\\

Stardew Valley & 13 & 14.43 & 13.08 & 139,625 & 16,092\\

\midrule Total & 394 & 223.36 & 197.85 & 2,072,084 & 287,232\\\bottomrule

\end{tabularx}\end{table}

\FloatBarrier

\subsection{Filtering and coverage recovery}

Of 11,891,441 screen candidates, 11,797,800 reach target checking after activity and causal-boundary checks. Removing 677,326 empty targets leaves 11,120,474 candidates; frame-retrieval checks remove another 1,826, leaving 11,118,648 legal dense samples. Sparse selection covers 10,732,300 of 10,738,981 eligible actions. Recovery places the remaining 6,681 actions in at least one target window. Coverage counts unique action ordinals; one action can occur in several overlapping targets.

\begin{figure}[!htbp]\centering

\includegraphics[width=\linewidth]{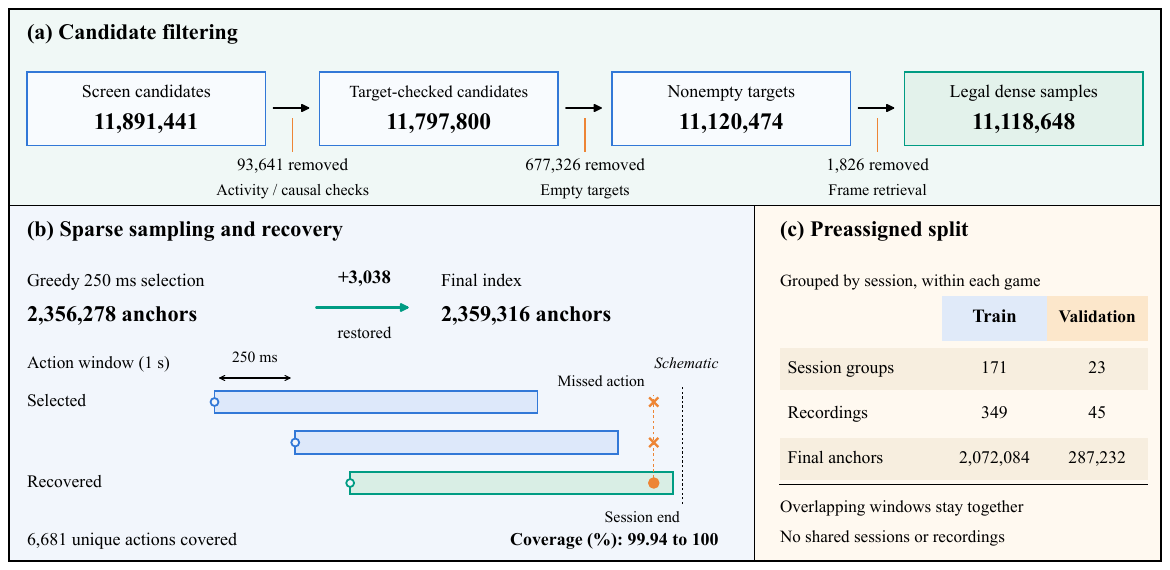}

\caption{Filtering, sparse anchor selection, coverage recovery, and the session-grouped split. All counts describe the same formal index.}\label{fig:sampling-coverage}\end{figure}

\FloatBarrier

\subsection{Recorded visual targets and timestamps}

\begin{figure}[!htbp]\centering

\includegraphics[width=.90\linewidth,height=.78\textheight,keepaspectratio]{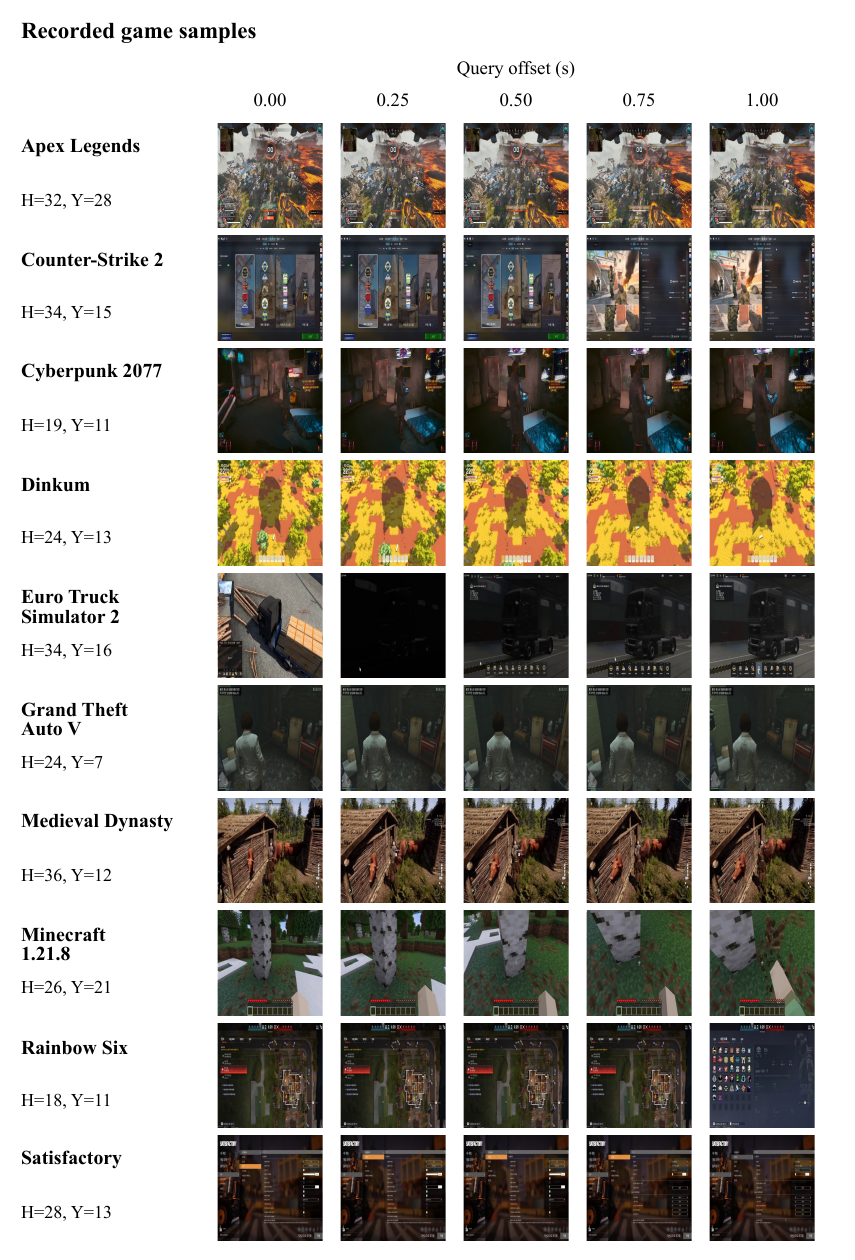}

\caption{Recorded five-frame training labels from ten games. $H$ and $Y$ denote the numbers of history and target events, respectively. The full subset has 20 games; this is a sample display, not an evaluation of generation quality or game generalization.}\label{fig:game-gallery}\end{figure}

Figure~\ref{fig:game-gallery} shows one train-index example from each of ten games, all using the formal one-second target. They illustrate the input distribution and label format, rather than compare model performance across games. Query offsets are $[0,250,500,750,1000]$ ms. Floor frame selection makes actual presentation timestamps approximately 0.34--16.53 ms earlier than requested in these examples. The accompanying data retain sample identifiers, action ordinals, and query and actual video timestamps.

\FloatBarrier

\section{Training, Data, and Checkpoint Details}\label{app:training}
\begin{table}[!htbp]
\caption{Desktop event-and-foresight pretraining configuration.}
\centering\small
\begin{tabularx}{\linewidth}{lY}\toprule
Setting & Value\\\midrule
Initialization & Expanded public InternVLA-A1.5-base\\
Human-event anchors & 2,072,084\\
Batch & 8 devices $\times$ 1 sample $\times$ 16 accumulation steps = 128\\
Optimizer & AdamW; learning rate $2\times10^{-5}$; betas $(.9,.95)$; epsilon $10^{-8}$\\
Regularization & Weight decay .01; gradient norm clipping 1.0\\
Schedule / seed & 200-update linear warmup, then constant; seed 42\\
Precision / distribution & BF16; DeepSpeed ZeRO-2\\
Context / foresight & Maximum prefix 8,192 positions; 50 query tokens\\
Vocabulary & 250,416 rows\\
Configured trainable parameters & 2,682,148,480\\
Video offsets & $[0,250,500,750,1000]$ ms\\
Video teacher & Frozen Wan2.2 TI2V-5B denoiser and VAE\\
Loss coefficients & Event 1.0; video 1.0\\
Transfer checkpoint & 12,400 updates; 1,587,200 sample presentations\\\bottomrule
\end{tabularx}
\end{table}

\paragraph{Sample construction.}
The materializer uses media timestamps to align events and frames. The condition image equals the anchor playback frame, and event spans are checked against the expanded tokenizer. Frame retrieval uses source presentation timestamps with floor selection. The resolved one-second video schedule determines the pretraining target. At the end of a robot episode, future video queries are clamped to the final frame.

\paragraph{Robot data consistency.}
Data preparation converts LeRobot v2.1 into v3.0 and constructs suite-specific training views. The four suites contain 1,693 episodes and 273,465 frames, with two $256\times256$ cameras and 20 Hz timestamps. Conversion checks verify actions, state, timestamps, and indices before and after conversion, as well as frame counts and sampled video access. Suite-specific robot-type keys keep normalization statistics separate. Strict video decoding surfaces decoding errors directly. Both initialization arms use the same converted data and the same single-suite training protocol.

\paragraph{Checkpoint conversion.}
The bridge reconstructs the native architecture from the expanded seed, consolidates the ZeRO-2 checkpoint, and overlays the saved trainable tensors while preserving the frozen seed components. It retains the expanded vocabulary, foresight queries, projection, and temporal metadata, and sets the native forward mode to robot. Reload and configuration checks verify that the resulting artifact can be consumed by the downstream model. This conversion changes the checkpoint packaging while retaining the trained parameter values.

\subsection{Downstream optimization and model identity}\label{app:robot-protocol}
All downstream runs follow the suite-specific configuration in Table~\ref{tab:robot-config}. Each policy is trained on a single LIBERO suite for 20,000 updates. Language supervision is drawn only from labels available in the corresponding robot batches; no additional language dataset or InternData-A1 replay is used.

\begin{table}[!htb]
\caption{Downstream robot recipe. Run-summary values describe the suite-specific 20,000-update setting. Optimizer and module settings specify the associated implementation recipe.}
\label{tab:robot-config}\centering\small
\begin{tabularx}{\linewidth}{p{.34\linewidth}Y}\toprule
Setting & Value\\\midrule
Training organization & One policy per suite and initialization\\
Budget and batch & 20,000 updates; two GPUs $\times$ 16 samples = 32\\
Learning-rate schedule & Peak $5\times10^{-5}$; endpoint $5\times10^{-6}$; 400 effective warmup updates\\
Optimizer preset & AdamW; betas $(.9,.95)$; epsilon $10^{-8}$; weight decay .01; clipping norm 1.0\\
Training seed / arithmetic & 42; BF16; gradient checkpointing enabled\\
Robot targets & Absolute actions; state tokens; FAST/language supervision; video weight 1\\
Native objective & $10\mathcal L_{\mathrm{action}}+\mathcal L_{\mathrm{language}}+\mathcal L_{\mathrm{video}}$\\
Trainable components & Visual encoder, multimodal backbone, native expert and action projections, foresight tokens and projections\\
Frozen components & Wan denoiser and VAE\\
Action prediction / execution & 50 predicted steps; eight executed before replanning\\
Saving and evaluated step & Every 5,000 updates; comparisons use step 20,000\\
Evaluation & Seed 7; 10 tasks $\times$ 50 episodes; 10 settling steps; $224\times224$ images\\\bottomrule
\end{tabularx}
\end{table}

\begin{figure}[!htb]
\centering
\includegraphics[width=\linewidth]{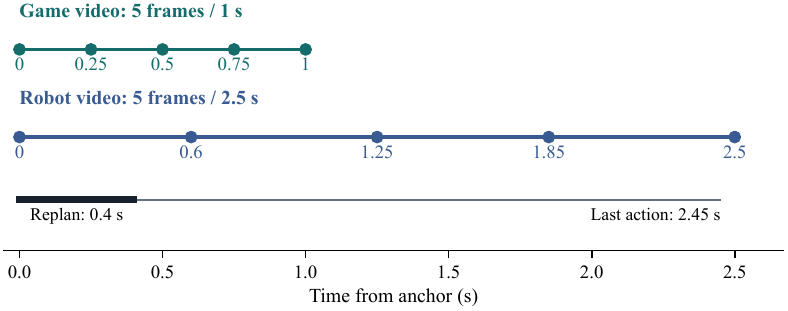}
\caption{Physical supervision times. Five-frame game targets span one second, whereas the robot video targets end at 2.50 seconds. The final sample in the 50-step robot action target is at 2.45 seconds; the separate replanning interval is eight executed actions, or 0.4 seconds.}
\label{fig:horizon}
\end{figure}

\FloatBarrier

For both initialization arms, the robot path uses the same local video-loss implementation: the condition latent is clean with timestep zero, whereas the future slice uses a sampled timestep. The public implementation uses a shared sampled timestep for the condition and future patches. Thus Base is the public \emph{initialization} fine-tuned with the shared local recipe; it is not the separately released, fully trained benchmark policy. Published scores obtained with different joint-training recipes are not directly comparable to this initialization contrast.

The transferred game checkpoint is step 12,400 and the compared downstream checkpoints are step 20,000. These identities specify the evaluated models without selecting a best score across a reported checkpoint sweep. Low-rank temporal adaptations appear only in the independent action-reconstruction diagnostics; they are not the robot fine-tuning procedure.

\FloatBarrier
\section{Complete LIBERO Task Results}\label{app:libero-tasks}

Each task is evaluated for 50 episodes per initialization. Tables give success counts and the game-minus-Base difference in percentage points. Task names retain the original benchmark wording. S/O/G/L identifiers map to the Spatial/Object/Goal/Long rows of the main-text heatmap.

\begin{table}[!htbp]\centering\footnotesize

\caption{Spatial: successes out of 50 episodes per task and arm.}\label{tab:tasks-spatial}

\begin{tabularx}{\linewidth}{lYrrr}\toprule

ID & Original task name & Base & Game & $\Delta$ pp\\\midrule

S01 & pick up the black bowl between the plate and the ramekin and place it on the plate & 42 & 49 & +14\\

S02 & pick up the black bowl next to the ramekin and place it on the plate & 47 & 47 & +0\\

S03 & pick up the black bowl from table center and place it on the plate & 49 & 49 & +0\\

S04 & pick up the black bowl next to the plate and place it on the plate & 49 & 47 & -4\\

S05 & pick up the black bowl on the wooden cabinet and place it on the plate & 39 & 47 & +16\\

S06 & pick up the black bowl on the cookie box and place it on the plate & 39 & 48 & +18\\

S07 & pick up the black bowl in the top drawer of the wooden cabinet and place it on the plate & 32 & 45 & +26\\

S08 & pick up the black bowl on the ramekin and place it on the plate & 36 & 41 & +10\\

S09 & pick up the black bowl next to the cookie box and place it on the plate & 42 & 43 & +2\\

S10 & pick up the black bowl on the stove and place it on the plate & 42 & 47 & +10\\

\bottomrule\end{tabularx}\end{table}

\FloatBarrier

\begin{table}[!htbp]\centering\footnotesize

\caption{Object: successes out of 50 episodes per task and arm.}\label{tab:tasks-object}

\begin{tabularx}{\linewidth}{lYrrr}\toprule

ID & Original task name & Base & Game & $\Delta$ pp\\\midrule

O01 & pick up the alphabet soup and place it in the basket & 45 & 49 & +8\\

O02 & pick up the cream cheese and place it in the basket & 50 & 50 & +0\\

O03 & pick up the salad dressing and place it in the basket & 45 & 50 & +10\\

O04 & pick up the bbq sauce and place it in the basket & 43 & 49 & +12\\

O05 & pick up the ketchup and place it in the basket & 43 & 49 & +12\\

O06 & pick up the tomato sauce and place it in the basket & 42 & 48 & +12\\

O07 & pick up the butter and place it in the basket & 48 & 50 & +4\\

O08 & pick up the milk and place it in the basket & 49 & 49 & +0\\

O09 & pick up the chocolate pudding and place it in the basket & 50 & 50 & +0\\

O10 & pick up the orange juice and place it in the basket & 45 & 50 & +10\\

\bottomrule\end{tabularx}\end{table}

\FloatBarrier

\begin{table}[!htbp]\centering\footnotesize

\caption{Goal: successes out of 50 episodes per task and arm.}\label{tab:tasks-goal}

\begin{tabularx}{\linewidth}{lYrrr}\toprule

ID & Original task name & Base & Game & $\Delta$ pp\\\midrule

G01 & open the middle drawer of the cabinet & 50 & 49 & -2\\

G02 & put the bowl on the stove & 48 & 46 & -4\\

G03 & put the wine bottle on top of the cabinet & 40 & 49 & +18\\

G04 & open the top drawer and put the bowl inside & 41 & 33 & -16\\

G05 & put the bowl on top of the cabinet & 41 & 48 & +14\\

G06 & push the plate to the front of the stove & 46 & 40 & -12\\

G07 & put the cream cheese in the bowl & 42 & 39 & -6\\

G08 & turn on the stove & 50 & 50 & +0\\

G09 & put the bowl on the plate & 47 & 45 & -4\\

G10 & put the wine bottle on the rack & 37 & 44 & +14\\

\bottomrule\end{tabularx}\end{table}

\FloatBarrier

\begin{table}[!htbp]\centering\footnotesize

\caption{Long: successes out of 50 episodes per task and arm.}\label{tab:tasks-long}

\begin{tabularx}{\linewidth}{lYrrr}\toprule

ID & Original task name & Base & Game & $\Delta$ pp\\\midrule

L01 & put both the alphabet soup and the tomato sauce in the basket & 42 & 24 & -36\\

L02 & put both the cream cheese box and the butter in the basket & 50 & 48 & -4\\

L03 & turn on the stove and put the moka pot on it & 48 & 50 & +4\\

L04 & put the black bowl in the bottom drawer of the cabinet and close it & 47 & 49 & +4\\

L05 & put the white mug on the left plate and put the yellow and white mug on the right plate & 14 & 30 & +32\\

L06 & pick up the book and place it in the back compartment of the caddy & 43 & 46 & +6\\

L07 & put the white mug on the plate and put the chocolate pudding to the right of the plate & 36 & 39 & +6\\

L08 & put both the alphabet soup and the cream cheese box in the basket & 35 & 47 & +24\\

L09 & put both moka pots on the stove & 44 & 34 & -20\\

L10 & put the yellow and white mug in the microwave and close it & 40 & 44 & +8\\

\bottomrule\end{tabularx}\end{table}

\FloatBarrier

\section{Evaluation Metrics and Aggregation}\label{app:metrics}
This section fixes the conventions used for two separate evaluations. The LIBERO results measure robot task success, whereas the game-action diagnostics in Appendix~\ref{app:diagnostics} measure reconstruction of recorded keyboard and mouse input; the latter checkpoints are not used to produce robot-pretraining labels.

\subsection{LIBERO aggregation}
Each LIBERO suite contains 10 tasks with 50 evaluation episodes per task. Let $n_{s,k}$ be the number of successes for task $k$ in suite $s$. The suite success rate is
\begin{equation}
\widehat p_s=\frac{1}{10}\sum_{k=1}^{10}\frac{n_{s,k}}{50}=\frac{\sum_{k=1}^{10}n_{s,k}}{500}.
\end{equation}
Because all four suites have the same evaluation budget, the overall rate is both the equal-weight mean of the suite rates and the pooled success rate,
\begin{equation}
\widehat p=\frac{1}{4}\sum_s\widehat p_s=\frac{\sum_s\sum_k n_{s,k}}{2000}.
\end{equation}
Differences are reported as game-pretrained initialization minus Base in percentage points. The overall gain is therefore $100(1811-1718)/2000=4.65$ percentage points. These values summarize the observed rollouts. Because each initialization has one training run per suite, they do not estimate variation across independent training seeds; the suite averages should be read together with all 40 task-level results in Appendix~\ref{app:libero-tasks}.

\subsection{Game-action diagnostic metrics}
At each scored context opportunity $b$, let $S_b$ and $\widehat S_b$ be the reference and predicted sets of held keyboard keys or mouse buttons. Held completeness is the fraction of opportunities at which both required held-state records appear in the expected order. State fidelity is measured by micro-F1,
\begin{equation}
F_1=\frac{2\sum_b|S_b\cap\widehat S_b|}{2\sum_b|S_b\cap\widehat S_b|+\sum_b|\widehat S_b\setminus S_b|+\sum_b|S_b\setminus\widehat S_b|}.
\label{eq:f1}
\end{equation}
A missing prediction for a nonempty reference state contributes false negatives, while an opportunity empty on both sides contributes no term. This state measure is distinct from exact matching of press/release events. In a separate six-game analysis of the public D2E checkpoint, exact matching over 50 ms bins that were nonempty on at least one side gave macro-average keyboard and mouse-button accuracies of 16.11\% and 26.15\%; adding bins empty on both sides as correct raised them to 69.78\% and 96.04\%. This comparison demonstrates sensitivity to the scoring convention; it does not identify the convention used for the originally published values, and empty-bin accuracy is not used to support the diagnostic claims below.

Mouse motion is scored by summing horizontal and vertical displacement within 50 ms bins and computing Pearson correlations $\rho_x$ and $\rho_y$ over bins containing a valid movement event in either sequence; bins empty on both sides are excluded. We use $\bar\rho=(\rho_x+\rho_y)/2$ when a single summary is needed. Correlation measures agreement in temporal variation, not absolute scale or exact event timing. Accepted mouse-event counts describe output density and are not, by themselves, fidelity measures; correlations for empty or degenerate sequences are undefined.

Teacher-forced validation loss and free-generation metrics also answer different questions. The former evaluates next-token prediction under reference histories, while held completeness, state F1, displacement correlation, and event count characterize generated action streams. Appendix~\ref{app:diagnostics} defines H-Base, T-Real, T-Unknown, T-Neutral, and No-Time. Gate-zero T-Real retains the LoRA learned jointly with its temporal module and is therefore distinct from H-Base.

\clearpage
\section{Game Action Representation Diagnostics}\label{app:diagnostics}
These supporting experiments study reconstruction of game actions. Their checkpoints are distinct from the human-event pretraining route used for robot transfer, and none of their predictions supplies the robot-pretraining labels.

\subsection{Diagnostic data and evaluation budget}
The temporal-residual experiments use 4,096 training rows, 512 validation rows, 500 optimizer updates, global batch 128, and learning rate $2\times10^{-5}$. This corresponds to $500\times128/4096=15.625$ nominal passes over the training rows. Training contexts contain 8,192 tokens and diagnostic inference contexts contain 2,048. Free-generation diagnostics use one Apex Legends validation sequence: a 60-second prefix for component interventions and a five-minute segment for the temporal-arm, gate, and timestamp-deletion comparisons. Keeping these sequence windows fixed makes the behavior of each intervention directly inspectable.

\subsection{Events, persistent state, and physical time}
Let a desktop recording be $\mathcal D=\{e_i\}_{i=1}^{N}$, where $e_i=(c_i,t_i,u_i)$ denotes event type, physical timestamp, and payload. Screen payloads contain images, while action payloads contain key identities, button states, or mouse displacement. Our held-state desktop base, \textbf{H-Base}, uses explicit \texttt{KBD\_HELD} and \texttt{MOUSE\_HELD} records. These state targets replace keyboard press/release and mouse-button transition supervision, distinguishing this base from the published D2E model. For timed events, the serializer represents
\begin{equation}
q_i=\left\lfloor t_i/(10\,\mathrm{ms})\right\rfloor\bmod1000
\label{eq:timestamp}
\end{equation}
with three digit tokens. Held records encode state at designated context opportunities and do not carry these three time digits. The inverse-dynamics offset is $\tau=100$ ms, applied once under the inference convention.

\subsection{Relative-time residuals and component interventions}
For token embedding $h_j$, relative-time identifier $r_j$, and eligibility mask $m_j$, we introduce
\begin{equation}
\widetilde h_j=h_j+m_j\tanh(g)\,A_\phi\!\left(\operatorname{LN}(E_\eta[r_j])\right).
\label{eq:residual}
\end{equation}
Here, $E_\eta$ is a time embedding, $A_\phi$ is a two-layer adapter with width 128, SiLU, and dropout, and $g$ is a learned scalar. Image placeholders and padding are excluded. Relative time is measured from the first screen in the active context using 10 ms buckets. Unknown values, padding, and offsets beyond 20.47 seconds receive distinct identifiers (Appendix~\ref{app:time-contract}). For timed actions, training and generation both reveal the time feature at the third digit, whose hidden state predicts the first payload token.

The implemented held-event boundary has a narrower asymmetry: training assigns the opportunity time at \texttt{EVENT\_START}, whereas cached generation assigns it once the following held-type token is known. This one-token feature-availability difference is part of the evaluated residual design. It provides an implementation-specific alternative explanation for instability; the component interventions do not isolate its causal contribution.

Three arms use the same trainable parameter topology and optimization budget. \textbf{T-Real} supplies actual relative times and retains time digits. \textbf{T-Unknown} maps nonpadding temporal inputs to the unknown identifier. \textbf{T-Neutral} retains the relative-time channel but replaces the input embeddings of time digits with a shared learned vector. Its digit targets remain supervised; identical digit tokens used for mouse displacement keep their original embeddings. Each arm trains attention LoRA with rank 8, scale 16, and dropout 0.05, together with the temporal modules, for 4,987,137 trainable parameters.

We probe the trained T-Real model by disabling the temporal residual, LoRA updates, or both. These post-training interventions measure functional dependence within the same co-adapted checkpoint. A gate sweep rescales the effective gate as $\tanh(g_\alpha)=\alpha\tanh(g)$ while leaving all other tensors unchanged. Component removal and gate scaling therefore retain the learned checkpoint as their common reference.

\subsection{Timestamp-free serialization}
\textbf{No-Time} removes the three timestamp digits following each timed event type. Event order, held-state payloads, and mouse-displacement and wheel digits remain. The base is frozen and attention LoRA is trained without a temporal adapter or learned time head. The parser assigns each action generated before the next screen the structural time $\max(0,t_{\mathrm{next}}-10\,\mathrm{ms})$, before applying the inverse-dynamics offset.

Timestamp deletion simplifies the output grammar while changing the information available for sub-frame reconstruction. If two recordings have identical retained observations, payloads, and order but different action times, a deterministic decoder must assign the same reconstructed time to both. Appendix~\ref{app:identifiability} formalizes this ambiguity. Deletion also changes the number of events that fit in a fixed token context, linking temporal representation to effective physical history.

\subsection{Separating structure and action fidelity}
We report held completeness, keyboard and mouse state F1, displacement correlations, and accepted mouse-event counts under the conventions in Appendix~\ref{app:metrics}. These quantities measure complementary properties: valid output structure, state accuracy, motion agreement, and output density. Held F1 uses raw state outputs, and an increase in event count alone is not a reconstruction error. The public-checkpoint examples in Appendix~\ref{app:cases} further illustrate why displacement correlation and event density must be interpreted separately.

\subsection{Low teacher-forced loss can coexist with structural failure}
\begin{figure}[!htb]
\centering
\includegraphics[width=\linewidth]{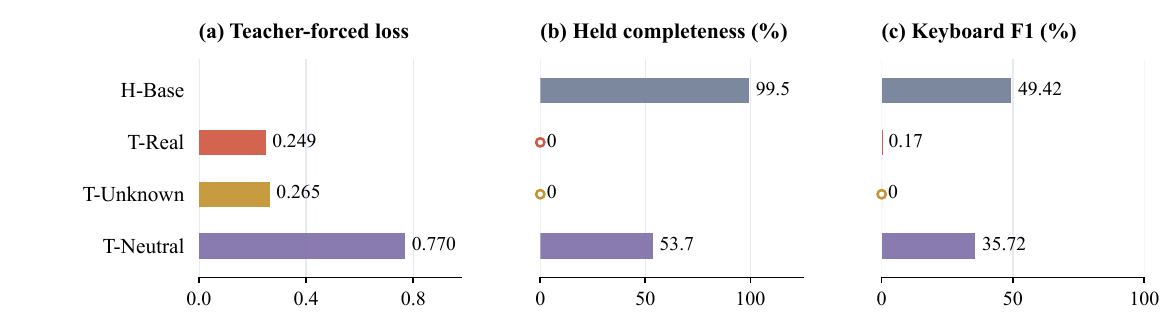}
\caption{Temporal representations separate teacher-forced loss from free-generation structure. Held completeness and keyboard F1 are plotted as percentages. H-Base supplies a generation reference; teacher-forced loss is compared only among the three adapted arms. Complete metrics, including mouse-state and motion results, appear in Table~\ref{tab:arms}.}
\label{fig:result-arms}
\end{figure}

Figure~\ref{fig:result-arms} compares the temporal-residual arms on the fixed five-minute sequence; Table~\ref{tab:arms} gives the detailed values. T-Real achieves the lowest teacher-forced loss among the three adaptations, yet held completeness falls from H-Base's 855/859 opportunities to zero. T-Unknown also loses held structure and generates no accepted mouse events. T-Neutral preserves 53.7\% completeness but produces only 43 mouse events, compared with 3,427 for H-Base. Thus the ranking induced by token loss does not describe the structure or action content of these autoregressive outputs.

\begin{figure}[!htb]
\centering
\includegraphics[width=\linewidth]{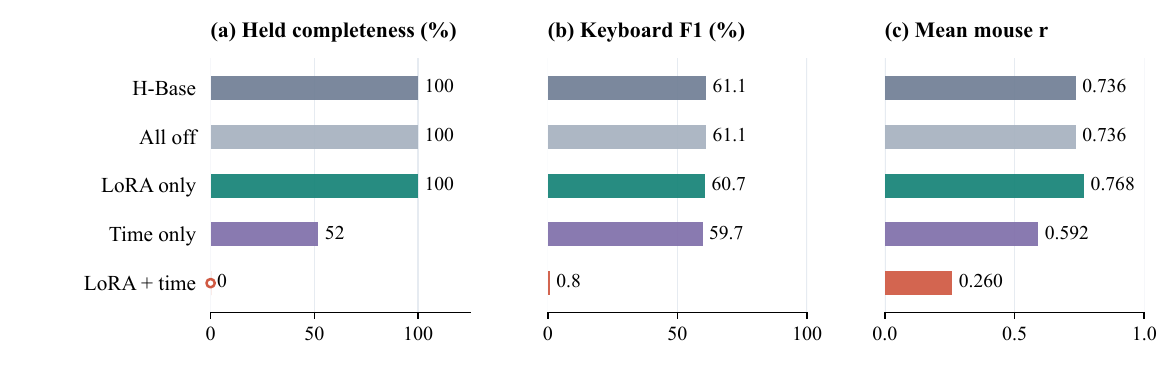}
\caption{Post-training component interventions on the 60-second prefix. Held completeness and keyboard F1 are plotted as percentages; mean mouse correlation averages the two displacement axes. ``All off'' reproduces H-Base on this prefix. These are interventions in one trained T-Real checkpoint, as specified in Table~\ref{tab:components}.}
\label{fig:result-components}
\end{figure}

Component interventions identify the functional dependence of this failure (Table~\ref{tab:components}; Figure~\ref{fig:result-components}). Disabling all additions reproduces H-Base's 759 action events exactly on the 60-second prefix. Keeping trained LoRA alone retains 100\% completeness. Keeping the temporal residual alone lowers completeness to 52\%, while enabling both components lowers it to zero. The enabled temporal pathway is therefore associated with disruption in this checkpoint, with a stronger failure when the co-trained LoRA is also enabled. T-Unknown's zero completeness shows that supplying real time values is not necessary for this failure. These observations identify functional dependence on a trained pathway, not the superiority of an independently trained LoRA-only model or a general defect of relative-time representations.

\begin{table}[!htb]
\caption{Post-training component interventions on the 60-second prefix. ``LoRA only'' and ``time only'' retain the corresponding components of the same trained T-Real checkpoint.}
\label{tab:components}\centering\small
\begin{tabular}{lrrrrr}\toprule
Configuration & $C$ (\%) & Kbd F1 & $\rho_x$ & $\rho_y$ & Events\\\midrule
H-Base & 100 & .611 & .723 & .750 & 759\\
All additions disabled & 100 & .611 & .723 & .750 & 759\\
LoRA only & 100 & .607 & .788 & .749 & 627\\
Time only & 52 & .597 & .724 & .460 & 592\\
LoRA + time & 0 & .008 & .316 & .204 & 351\\\bottomrule
\end{tabular}
\end{table}

\begin{figure}[!htb]
\centering
\includegraphics[width=\linewidth]{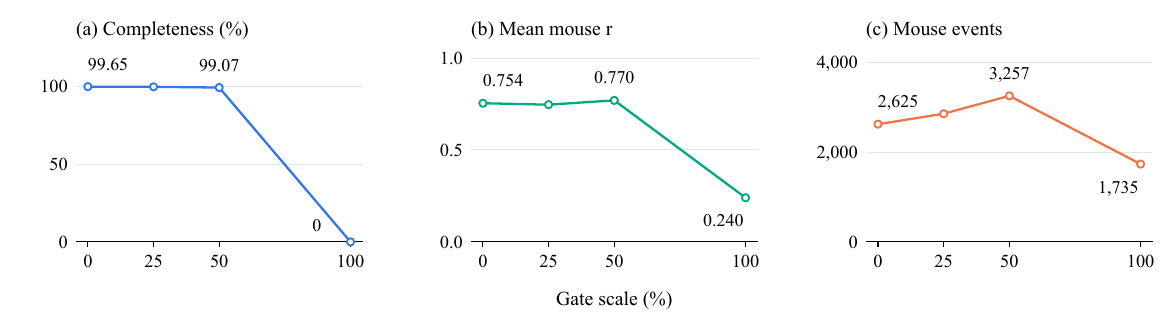}
\caption{Temporal-gate scan on the five-minute sequence. The horizontal axis shows $100\alpha\%$, where $\alpha$ is the scale in Table~\ref{tab:gate}; connecting lines are visual guides between the four evaluated settings. Mean mouse correlation averages the two displacement axes. Full gate strength collapses held structure, while moderate scaling changes both motion agreement and event density.}
\label{fig:result-gate}
\end{figure}

The gate sweep reveals joint changes in displacement agreement and output density (Figure~\ref{fig:result-gate}; Table~\ref{tab:gate}). At scale 0.5, $\rho_x$ and $\rho_y$ exceed the zero-gate values, while accepted events increase by 24.1\%. At full scale, held structure collapses. These joint changes make gate selection a multi-metric decision: improving one correlation does not guarantee stable event generation or appropriate event density.

\subsection{Grammatical validity and action fidelity separate after time deletion}
\begin{figure}[!htb]
\centering
\includegraphics[width=\linewidth]{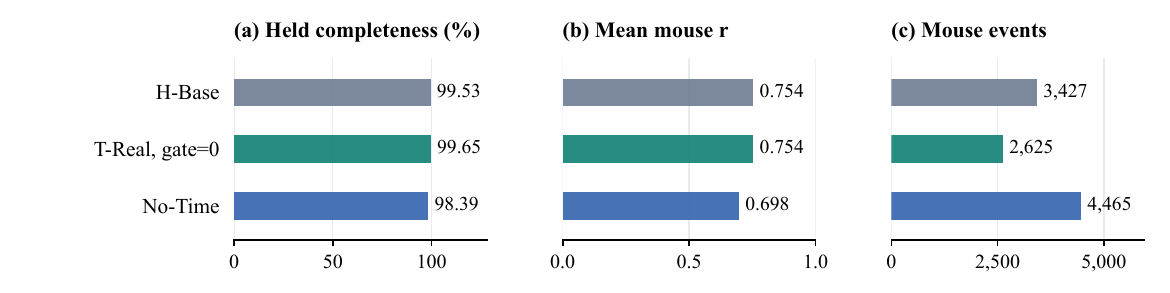}
\caption{Timestamp deletion retains high held completeness while reducing mean displacement correlation and increasing event count. The figure shows these three summaries on the same five-minute sequence; keyboard and mouse state F1 appear in Table~\ref{tab:notime}. The zero-gate reference retains T-Real's trained LoRA.}
\label{fig:result-notime}
\end{figure}

No-Time reaches 98.39\% held completeness (Figure~\ref{fig:result-notime}; Table~\ref{tab:notime}). Relative to H-Base, keyboard F1 decreases by 0.0309, mouse F1 by 0.0431, and mean displacement correlation by 0.0566. Meanwhile, 4,465 accepted mouse events represent a 30.3\% increase over H-Base and a 70.1\% increase over the zero-gate T-Real intervention. These results demonstrate that a valid event grammar can coexist with poorer action reconstruction and a substantial change in event density.

The comparison evaluates timestamp deletion together with its structural time assignment and changed token-context occupancy. The measured effect therefore belongs to the complete resulting event system, rather than an isolated token operation.

\FloatBarrier

\section{Implementation and Representation Details}
Table~\ref{tab:positioning} in the main text gives the method attribution. The following analyses concern implementation and independent game-action diagnostics.

\subsection{Time features at each token position}\label{app:time-contract}
A timed event has the token layout
\begin{equation*}
[\mathtt{START},\mathtt{TYPE},d_1,d_2,d_3,\mathtt{PAYLOAD},\mathtt{END}].
\end{equation*}
An input feature at position $j$ conditions prediction of token $j+1$. Relative buckets use the first screen of the current context as origin. Bucket $b\in[0,2047]$ maps to ID $3+b$; IDs 0, 1, and 2 denote padding, unknown, and overflow. The embedding therefore contains 2,051 rows. Absolute media timestamps anchor the ten-second serialization cycle; action timestamps are unwrapped monotonically. Ambiguous old packed rows with an action inside a screen gap of at least one complete cycle are rejected by strict reconstruction.

\begin{table}[!htbp]
\caption{Time-feature availability in the supplied implementation. ``Known'' means the event or opportunity time relative to the active screen anchor. Cached generation refers to a newly generated event.}
\label{tab:availability}\centering\small
\begin{tabularx}{\linewidth}{Ycc}\toprule
Input position & Teacher forcing & Cached generation\\\midrule
Timed action: START, TYPE, $d_1,d_2$ & Unknown & Unknown\\
Timed action: $d_3$, payload, END & Known & Known\\
Held event: START & Known & Unknown\\
Held event: held TYPE, payload, END & Known & Known\\\bottomrule
\end{tabularx}
\end{table}
Observed screens have externally available times. Completed held records in the context are annotated from their known opportunity times. For a new held event, the inference function identifies its class after the type token, whereas training uses the completed event span. This accounts for the START-row difference in Table~\ref{tab:availability}. A source-function prefix check reproduces the difference without loading a learned model. It establishes the feature convention, not an explanation of the measured failure.

The residual eligibility mask excludes padding and image placeholders; it does not suppress the learned unknown-time embedding. T-Unknown replaces every nonpadding time ID with unknown, so it retains an active residual despite removing real time values. T-Neutral changes only the embeddings at the three timestamp-digit positions; identical digits used in mouse payloads retain their ordinary embeddings.

\subsection{Diagnostic data, optimization, and decoding}
Pilot construction preserves the source train/validation session assignment. It takes a contiguous prefix from the first sufficiently long split-owned session per game, packs each session separately, drops artificial episode tails, and selects approximately equal packed-row quotas across games: 4,096 rows from 20 training games and 512 rows from 10 validation games. No test split is used for the reported pilot or gate choices. Adjacent rows within a session and repeated optimization passes are not independent recordings.

\begin{table}[!htbp]
\caption{Configuration of the evaluated desktop adaptations. The residual-arm validation loss is aggregated over 264,276 supervised shifted tokens in the 512 validation rows.}
\label{tab:desktop-config}\centering\small
\begin{tabularx}{\linewidth}{lYY}\toprule
Setting & T-Real / T-Unknown / T-Neutral & No-Time pilot\\\midrule
Train / validation rows & 4,096 / 512 & 4,096 / 512\\
Optimizer updates / global batch & 500 / 128 & 500 / 128\\
Learning rate & $2\times10^{-5}$ & $2\times10^{-5}$\\
Train / inference context & 8,192 / 2,048 tokens & 8,192 / 2,048 tokens\\
Attention LoRA & Rank 8, scale 16, dropout .05 & Rank 8, scale 16, dropout .05\\
Trainable parameters & 4,987,137 & 1,081,344\\
Temporal adapter / gate & Width 128; initial scalar .001 & None\\
Nominal training passes & 15.625 & 15.625\\\bottomrule
\end{tabularx}
\end{table}
The three residual arms use the same initialization and training seed, 8 A100 devices, BF16, cosine learning-rate decay, and 15 warmup updates. Their stored final gate scalars are .006134, .002426, and .004578 for T-Real, T-Unknown, and T-Neutral. These scalars multiply learned features; their magnitude alone is not a residual-to-embedding norm ratio. Component removal operates after this joint training and leaves the retained tensors unchanged.

Residual-arm inference selects each next token by argmax, carries a causal cache within the event, and stops at EVENT\_END, SCREEN, or the configured token cap. The held metrics in this paper use raw outputs. A separate smoother uses a 400 ms maximum adjacent gap and three agreeing records on each side, and does not define the reported raw F1. The inverse-dynamics shift is applied once when output timestamps are written. Completeness and state F1 have distinct eligibility rules: the five-minute diagnostic contains 859 context records for completeness, with one insufficient-context record excluded from state scoring. The 60-second component diagnostic contains 173 opportunities.

\subsection{Interpreting loss and supervising video}\label{app:loss-conventions}
For the residual-arm validation objective, partition supervised tokens into disjoint categories $c$ (such as type, timestamp, and payload). The exact averaging identity is
\begin{equation*}
\mathcal L_{\mathrm{val}}=\sum_c\frac{N_c}{N}\mathcal L_c.
\end{equation*}
This explains why the aggregate objective and a particular structural event can move differently; it is not a measurement of category-specific losses. The causal desktop pretraining objective in Equation~\ref{eq:nep} instead averages within each sample before averaging across samples. These reductions refer to different experiments.

In the video objective, the clean clip latent is $48\times2\times14\times14$ per sample and the independently encoded condition is $48\times1\times14\times14$. The condition slice contributes zero error. Dividing by all 18,816 latent elements therefore equals half the mean error over the 9,408 future elements. A timestep index is sampled uniformly from $\{0,\ldots,999\}$, and its noise level is read from the shift-5.0 scheduler; this is distinct from sampling the noise level uniformly. Frozen denoiser parameters still permit derivatives with respect to the projected foresight features. Both downstream initialization arms use this same implementation.

\FloatBarrier
\section{Existing Qualitative Examples and Metric Interpretation}\label{app:cases}
The following examples come from the evaluation of the public D2E checkpoint supplied with this study. They illustrate metric behavior and are separate from H-Base and the temporal-adaptation checkpoints. Each table row is one five-second window, not an independent repetition of a temporal intervention.

\begin{figure}[!htbp]
\centering
\includegraphics[width=\linewidth]{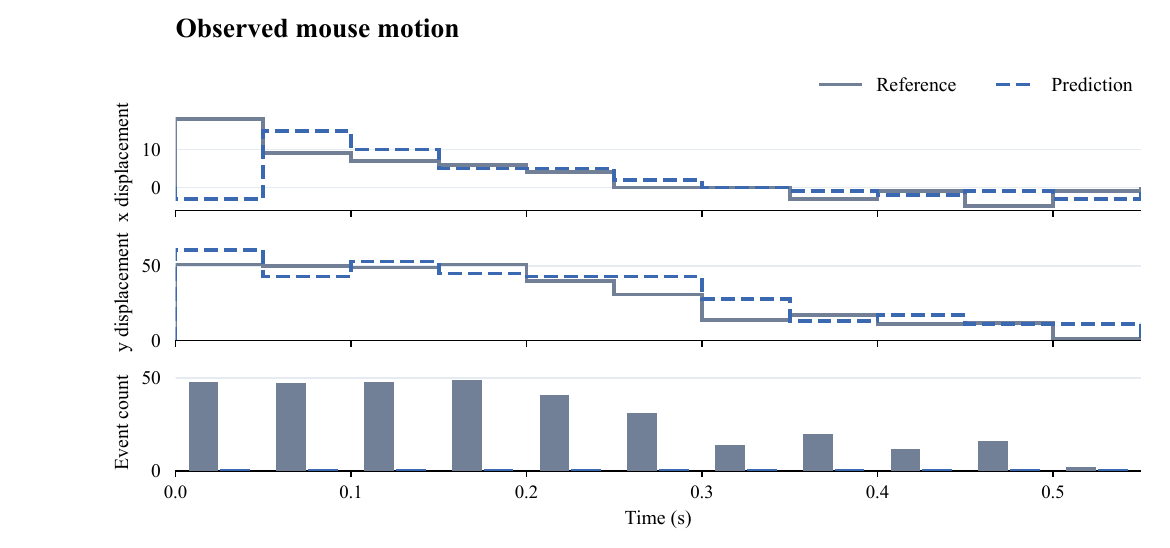}
\caption{The first 0.55 seconds of the Brotato example from the public D2E checkpoint. The upper panels sum reference and predicted $x/y$ displacement within 50 ms bins; the lower panel counts raw events in the same bins (328 reference and 11 predicted). The example illustrates the distinction between accumulated motion and event density, not a temporal-adaptation effect.}
\label{fig:existing-case}
\end{figure}

\begin{table}[!htbp]
\caption{Existing mouse-motion examples. Correlations are computed from accumulated displacement in 50 ms bins; counts describe raw reference and predicted events over the complete five-second window.}
\label{tab:existing-cases}\centering\small
\begin{tabular}{lrrrr}\toprule
Game & $\rho_x$ & $\rho_y$ & Reference events & Predicted events\\\midrule
Brotato & .9932 & .9924 & 1,535 & 55\\
GTA V & .9924 & .9923 & 1,794 & 60\\
Apex Legends & $-.3300$ & .4019 & 349 & 85\\\bottomrule
\end{tabular}
\end{table}
The corresponding recording windows are 26:12--26:17, 52:48--52:53, and 09:49--09:54. Figure~\ref{fig:existing-case} shows the first eleven consecutive bins of the Brotato window. Its plotted interval is 0.55 seconds; the correlations in Table~\ref{tab:existing-cases} refer to the complete five seconds. The plotting values are transcribed from the original per-bin table without filling unobserved bins or interpolating a trajectory.

Mouse-displacement scoring uses events with zero button flags, includes a bin when either sequence contains a valid movement event, and excludes bins empty on both sides. Correlation measures agreement in variation, not absolute scale or timestamp accuracy. Consequently, the No-Time count increase in Appendix~\ref{app:diagnostics} is described as a density change. Its lower state F1 and displacement correlation provide separate evidence about action reconstruction under the implemented parser.
\FloatBarrier

\section{Temporal Identifiability}\label{app:identifiability}
The following observation formalizes the information loss that can arise when sub-frame timestamps are removed while screens, payloads, and event order are retained.
\paragraph{Observation: sub-frame ambiguity after timestamp deletion.}
Suppose two admissible recordings have identical observed screens, event payloads, and event order after timestamp deletion, but one corresponding action has times $t$ and $t'$ with $|t-t'|=d$. Any deterministic reconstruction using only this retained representation incurs absolute time error at least $d/2$ on one of the two recordings.

\begin{proof}
Identical retained inputs give the same reconstruction $\widehat t$. By the triangle inequality, $d\leq|t-\widehat t|+|\widehat t-t'|$, so at least one term is at least $d/2$.
\end{proof}
The condition concerns observationally indistinguishable inputs. Visual evidence can still resolve timing in other cases. The practical implication is that successful parsing alone does not determine how much physical-time information the representation preserves. In our No-Time configuration, assigning events relative to the next screen gives a defined structural schedule whose fidelity is evaluated through the resulting action stream.

\clearpage
\section{Detailed Result Tables}\label{app:result-tables}
The tables below provide the numerical values underlying the three diagnostic result figures in Appendix~\ref{app:diagnostics}. Metric definitions are given in Appendix~\ref{app:metrics}, and the model variants are defined in Appendix~\ref{app:diagnostics}.

\begin{table}[!htbp]
\caption{Temporal-residual adaptations on the five-minute Apex diagnostic sequence. $C$ is held completeness. Loss compares the three trained arms on the same 512 validation rows; the H-Base row provides a generation reference. ``Undef.'' denotes undefined correlation.}
\label{tab:arms}\centering\small
\begin{tabular}{lrrrrrr}\toprule
Model & Loss & $C$ (\%) & Kbd F1 & Mouse F1 & $\rho_x/\rho_y$ & Events\\\midrule
H-Base & -- & 99.5 & .4942 & .2121 & .7662/.7423 & 3,427\\
T-Real & .24934 & 0.0 & .0017 & .0000 & .2823/.1977 & 1,735\\
T-Unknown & .26507 & 0.0 & .0000 & .0000 & Undef. & 0\\
T-Neutral & .77015 & 53.7 & .3572 & .0448 & .0173/.1159 & 43\\\bottomrule
\end{tabular}
\end{table}

\begin{table}[!htbp]
\caption{Effective temporal-gate scaling on the five-minute diagnostic sequence. Scale zero retains the LoRA learned jointly with the temporal module.}
\label{tab:gate}\centering\small
\begin{tabular}{rrrrrrr}\toprule
Scale & $C$ (\%) & Kbd F1 & Mouse F1 & $\rho_x$ & $\rho_y$ & Events\\\midrule
0.00 & 99.65 & .4891 & .1875 & .7918 & .7167 & 2,625\\
0.25 & 99.53 & .4907 & .1875 & .7860 & .7068 & 2,858\\
0.50 & 99.07 & .4893 & .2258 & .8026 & .7372 & 3,257\\
1.00 & 0.00 & .0017 & .0000 & .2823 & .1977 & 1,735\\\bottomrule
\end{tabular}
\end{table}

\begin{table}[!htbp]
\caption{Timestamp-deletion comparison on the five-minute diagnostic sequence. The gate-zero reference retains T-Real's jointly trained LoRA. Baseline entries and deltas are specific to this comparison; numerical details are given in Appendix~\ref{app:metrics}.}
\label{tab:notime}\centering\small
\begin{tabular}{lrrrrrr}\toprule
Model & $C$ (\%) & Kbd F1 & Mouse F1 & $\rho_x$ & $\rho_y$ & Events\\\midrule
H-Base & 99.53 & .4942 & .2121 & .7662 & .7423 & 3,427\\
T-Real, gate=0 & 99.65 & .4891 & .1875 & .7918 & .7167 & 2,625\\
No-Time & 98.39 & .4633 & .1690 & .6839 & .7114 & 4,465\\\bottomrule
\end{tabular}
\end{table}

\FloatBarrier

\end{document}